# Extracting and analyzing 3D histomorphometric features related to perineural and lymphovascular invasion in prostate cancer

Sarah S.L. Chow, Rui Wang*, Robert B. Serafin*, Yujie Zhao*, Elena Baraznenok, Xavier Farré, Jennifer Salguero-Lopez, Gan Gao, Huai-Ching Hsieh, Lawrence D. True, Priti Lal, Anant Madabhushi, and Jonathan T.C. Liu

*Abstract*— Diagnostic grading of prostate cancer (PCa) relies on the examination of 2D histology sections. However, the limited sampling of specimens afforded by 2D histopathology, and ambiguities when viewing 2D cross-sections, can lead to suboptimal treatment decisions. Recent studies have shown that 3D histomorphometric analysis of glands and nuclei can improve PCa risk assessment compared to analogous 2D features. Here, we expand on these efforts by developing an analytical pipeline to extract 3D features related to perineural invasion (PNI) and lymphovascular invasion (LVI), which correlate with poor prognosis for a variety of cancers. A 3D segmentation model (nnU-Net) was trained to segment nerves and vessels in 3D datasets of archived prostatectomy specimens that were optically cleared, labeled with a fluorescent analog of H&E, and imaged with open-top light-sheet (OTLS) microscopy. PNI- and LVI-related features, including metrics describing cancer-nerve and cancer-vessel proximity, were then extracted based on the 3D nerve/vessel segmentation masks in conjunction with 3D masks of cancer-enriched regions. As a preliminary exploration of the prognostic value of these features, we trained a supervised machine learning classifier to predict 5-year biochemical recurrence (BCR) outcomes, finding that 3D PNI-related features are moderately prognostic and outperform 2D PNI-related features (AUC = 0.71 vs. 0.52). Source code is available at https://github.com/sarahrahsl/SegCIA.git.

## I. INTRODUCTION

PROSTATE cancer (PCa) is the most diagnosed cancer among men and remains the second leading cause of cancer-related deaths in men in the United States [1]. Currently, the gold standard for diagnosing and grading prostate cancer relies on the examination of hematoxylin and eosin (H&E)- stained 2D tissue sections (around 5-µm thick) obtained from formalin-fixed paraffin-embedded (FFPE) tissues. If prostate cancer is observed, then the aggressiveness of the cancer is determined based on the visual interpretation of the spatial architecture of the prostate glands according to the Gleason grading system [2]–[4]. One challenge with the current histology workflow is that it captures ~1% of the total biopsy volume, which can misrepresent heterogeneous tissue structures[5], [6]. Moreover, interpreting complex 3D biological structures can be ambiguous and misleading when viewing 2D histology sections [7]–[9]. These limitations contribute to substantial inter-observer variability among pathologists, particularly when grading low- to intermediate-grade cancers [10]–[12]. As a result, patients may be under-treated, risking metastasis and death [13], or over-treated, leading to unnecessary side effects such as impotence and incontinence [14].

Recent advances in high-throughput 3D microscopy, in conjunction with tissue-clearing and rapid fluorescence-labeling techniques [15], [16], have enabled the development of non-destructive 3D pathology. Slide-free 3D pathology offers several advantages over conventional 2D pathology, including: (1) comprehensive imaging of large clinical specimens, as opposed to the limited sampling achieved through standard thin tissue sections, and (2) volumetric visualization and unambiguous characterization of diagnostically important 3D structures. These advantages have been shown to improve diagnostic assessments by both human observers and

Research reported in this publication was supported by the National Cancer Institute (NCI) under R01CA268207 (Liu and Madabhushi); the Department of Defense (DoD) Prostate Cancer Research Program (PCRP) under W81XWH-18-10358 (Liu and True) and W81XWH-20-1-0851 (Madabhushi and Liu); the Pacific Northwest Prostate Cancer SPORE P50CA97186 (True), and the Canary Foundation.

Sarah S.L Chow, Rui Wang, Yujie, Huai-Ching Hsieh, and Jonathan T.C. Liu are with the Department of Pathology, Stanford University, Stanford, CA 94305 USA, and with University of Washington, Seattle, WA 98195 USA. Jonathan T.C. Liu and Huai-Ching Hsieh are also with Bioengineering, Stanford University, Stanford, CA 94305 USA (e-mail: slschow@stanford.edu, rwang98@uw.edu, yujiez@stanford.edu, hchsieh@stanford.edu, jonliu@stanford.edu).

Robert B. Serafin, Elena Baraznenok, were with University of Washington, Seattle, WA 98195 USA. Rob Serafin is now with University of Chicago, Chicago, IL 60637 USA. Elena Baraznenok is now with the UCB-UCSF Joint Graduate Program in Bioengineering, University of California, Berkeley, CA 94720 USA (e-mail: Robert.b.serafin@gmail.com, ebaraznenok@berkeley.edu).

Lawrence D. True, and Gan Gao are with University of Washington, Seattle, WA98195, USA (e-mail: ltrue@uw.edu, gangao@uw.edu).

Xavier Farre is with Public Health Agency of Catalonia, Lleida, Spain (e-mail: xfarre@fullbrightmail.org).

Jennifer S. Lopez and Anant Madabhushi are with Wallace H. Coulter Department of Biomedical Engineering at Georgia Institute of Technology and Emory University, Atlanta, GA 30322 USA. Anant Madabhushi is also with the Atlanta VA Medical Center, Atlanta, GA 30033, USA (e-mail: Jennifer.salguero.lopez@emory.edu, anantm@emory.edu).

Priti Lal is with the Department of Pathology and Laboratory Medicine in University of Pennsylvania, Philadelphia, PA 19104 USA (e-mail: pritilal@pennmedicine.upenn.edu).

*Contributed equally

computational algorithms [9], [17]–[19]. Amongst various techniques for 3D pathology [18], [20]–[22], open-top light-sheet (OTLS) microscopy [7], [8], [23]–[25] is able to generate high-quality 3D pathology datasets of large clinical specimens at spatial resolutions comparable to conventional H&E histology. Without the need for physical sectioning, large clinical specimens can be rapidly labelled with small-molecule (quickly diffusing) fluorescent analogs of H&E, optically cleared with gentle and reversible protocols, and imaged intact and *in toto* [24], [25].

A challenge with 3D pathology is the size of the data – a single biopsy can generate on the order of 10 to 100 GB of high-resolution 3D microscopy data (uncompressed). This necessitates the development of scalable methods for extracting meaningful features and insights. In recent years, a number of computational methods [18] have been developed to analyze 3D pathology datasets. While end-to-end deep learning models have shown impressive performance for tasks such as risk classification, these fully automated methods often lack interpretability and require extensive validation before they can be clinically deployed. In contrast, there is great interest in designing interpretable classifiers based on intuitive hand-crafted features derived from 3D structures/primitives. These explainable methods offer physical insights and can aid in both hypothesis generation and testing for biologists and pathologists. Recent studies on hand-crafted 3D glandular features [19], including volume ratios, tortuosity, and curvature, and similar studies on shape-based nuclear features [17], have shown superior performance for predicting patient outcomes compared to corresponding 2D features.

In addition to glandular morphology, which is the basis for the Gleason grading system, and nuclear morphology, whose structural alterations can reflect underlying molecular changes within diseases, modes of tumor invasion such as perineural invasion (PNI) and lymphovascular invasion (LVI) are common indicators of aggressive disease. PNI and LVI are characterized by tumor cells infiltrating along nerves or entering blood and lymphatic vessels, respectively, providing routes for metastasis. The presence of PNI or LVI has been associated with poor prognosis in prostate cancer [26]–[28] and PNI especially can be used to guide treatment decisions for prostate cancer patients [29].

Despite their clinical significance, screening for PNI and LVI is not a standardized component of histopathological evaluation of prostate cancers [30]. This may be because PNI and LVI are rare in thin histology sections, which makes it challenging to identify PNI / LVI in an objective, quantifiable, and reproducible manner. Although immunohistochemical staining for nerves or vessels can facilitate the identification of PNI or LVI, it requires additional time, reagents and labor, and therefore is not routinely implemented in clinical workflows. With the advent of 3D pathology, we hypothesize that a volumetric assessment of the interaction between cancer regions and neural or vascular structures would enable more accurate and reliable evaluation of PNI and LVI, offering diagnostic insights that are often missed in conventional 2D histology. The key contributions of this study are as follows:

- We present a deep learning-based pipeline for nerve and vessel segmentation in 3D prostate pathology datasets (H&E-analog staining) using ground-truth immunofluorescence datasets for supervised training.
- We extract novel interpretable features related to perineural invasion (PNI) and lymphovascular invasion (LVI) from the 3D segmentation masks of nerves, vessels, and cancer-enriched regions.
- We provide a preliminary exploration of the prognostic value of our extracted features for the prediction of 5-year biochemical recurrence (BCR) outcomes for low- to intermediate-risk prostate cancer patients.

## II. Related work

### A. Nerve and vessel segmentation in 2D H&E histopathology images

Accurate segmentation of histological structures is essential for quantitative morphometric analysis, as such segmentations enable the extraction of biologically meaningful features that inform cancer risk assessment. However, segmenting nerves and small vessels from H&E pathology images is challenging because these structures are convoluted branching structures with unpredictable geometries. They are also often subtle and ambiguous in H&E images. For example, peripheral nerves may appear as pale, eosinophilic fascicles with indistinct borders, while small vessels or capillaries may appear collapsed or tangentially sectioned. These morphological ambiguities contribute to low interobserver agreement in the identification of LVI [31].

To accurately segment nerves and vessels in biological tissues, deep learning-based automated segmentation approaches have been developed. However, most existing algorithms only operate on thin 2D histology images [32]–[34], or on 3D microscopy datasets of animal models in which vessels or neurons are labeled with genetic reporters or systemic agents [35]. As such, these methods do not generalize to thick, volumetric datasets labeled with H&E-analog stains. In this work, we introduce a direct 3D segmentation method for nerves and vessels in volumetric pathology datasets, enabling unprecedented large-scale analysis of PNI and LVI in 3D datasets of human tissues stained with a fluorescent analog of H&E.

### B. Deep-learning-based segmentation in 3D pathology datasets

The segmentation of tissue structures in 3D pathology datasets is typically achieved through deep-learning (DL)–based methods or traditional computer-vision (CV) approaches. DL-based segmentation requires large, manually annotated training datasets, which are time-consuming and labor-intensive to produce, whereas CV methods often require structure-specific immunolabeling, which can be costly and slow due to the limited diffusion of antibodies through large tissue volumes. To address these limitations, Xie *et*

*al.* developed image-translation-assisted segmentation in 3D (ITAS3D) [19], in which H&E-analog input datasets are computationally converted into synthetic immunofluorescence outputs, which in turn facilitate 3D segmentation of the immunolabeled structures via CV methods. This annotation-free segmentation approach leverages rapid and inexpensive small molecule staining (i.e. H&E-analog staining), circumventing the cost and time of performing 3D immunolabeling. However, ITAS3D involves a two-step procedure combining DL-based image translation and CV segmentation, the latter of which still requires manual parameter tuning. In contrast, our method trains a direct 3D segmentation model using immunolabeling-derived images to form ground truth segmentation masks (for training), thus enabling a one-step segmentation of nerves and vessels from H&E-analog datasets.

*C.  Segmentation masks of cancer-enriched regions*

To analyze the spatial relationship between cancer content surrounding neural or vascular structures, the cancer-enriched regions must first be localized within the tissue. The SIGHT pipeline [36] —a deep-learning–based 3D image translation framework—provides a practical solution for this task. SIGHT uses the ITAS3D framework to synthetically label two cytokeratin markers that are differentially expressed in malignant versus benign prostate glands. These synthetic labels are then used to generate continuous-valued heatmaps that highlight cancer-enriched regions within the tissue. By applying a fixed threshold, these heatmaps are binarized to produce 3D segmentation masks of cancer-enriched regions at glandular resolution. By overlaying these cancer masks with nerve or vessel segmentation masks, we compute features that capture the spatial interaction between cancer-enriched tissue regions and adjacent neural or vascular structures, serving as a proxy for PNI and LVI.

## III.  Materials and Methods

In this study, we developed a computational pipeline to quantify three-dimensional (3D) features associated with perineural invasion (PNI) and lymphovascular invasion (LVI) from volumetric prostate pathology datasets. The pipeline involves training a 3D deep-learning model (nnU-Net) to segment nerves and vessels from optically cleared prostate specimens labeled with a fluorescent analog of hematoxylin and eosin (H&E) and imaged using open-top light-sheet (OTLS) microscopy, thereby eliminating the need for additional immunolabeling once the segmentation model is trained (**Fig 1**). The resulting nerve- and vessel-segmentation masks are integrated with previously generated cancer-enriched region masks to extract 3D spatial features that characterize tumor–nerve and tumor–vessel proximity/interactions. As a preliminary investigation, we apply this pipeline to a cohort of 120 prostate biopsies to evaluate the prognostic value of PNI- and LVI-related 3D features.

The following subsections describe each component of the workflow in detail, including tissue preparation (Section III-A), OTLS imaging (Section III-B), deep-learning–based 3D segmentation (Section III-C), segmentation evaluation (Section III-D), feature extraction (Section III-E), and feature evaluation (Section III-F). This integrated framework provides the foundation for future studies investigating PNI, LVI, and related processes in large-scale 3D pathology datasets.

*A.  Datasets and processing of archived tissue for training and validation*

For training, we collected archived formalin-fixed, paraffin-embedded (FFPE) blocks from 15 radical prostatectomy (RP) specimens archived in an IRB-approved genitourinary biorepository at the University of Washington (Seattle, WA; Study 00004980). These specimens were obtained from patients originally diagnosed with low- to intermediate-risk PCa (Gleason grade groups 1-3). Each FFPE block was sectioned into 100-μm thick slices (tissue curls), and tri-labeled with a fluorescent analog of H&E plus an antibody of interest, according to a protocol adapted from the iDISCO method [15] and modified by Xie et al. [19]. Cytoplasmic staining was achieved with Alexa Fluor 488 NHS Ester (5 μg/mL), and nuclear staining with SYTO 85 (5 μM). Immunolabeling was performed using either a mouse anti-human PGP9.5 (ab8189, Abcam, USA) or anti-human CD31 (M082329-2, Agilent, USA) primary antibody (1:100 dilution), followed by Alexa 647-conjugated secondary antibody (715-605-150, Jackson ImmunoResearch, USA) (1:100 dilution). After staining, the samples were dehydrated and optically cleared in ethyl cinnamate before imaging.

For validation, archived FFPE prostatectomy specimens were used from 120 patients with known clinical outcomes from the University of Pennsylvania. A genitourinary pathologist (P.L.) reviewed each case to identify a region most representative of the tumor, from which one 3-mm diameter punch biopsy (average thickness ∼0.5-mm) was extracted per case. Each punch biopsy was labeled with a small-molecule fluorescent analog of H&E consisting of TO-PRO-3 Iodide (1:500 dilution) for nuclear staining and Alcoholic Eosin Y (1:100 dilution) for cytoplasmic staining, according to a previously published protocol [24]. Stained tissues were then dehydrated and cleared in ethyl cinnamate for imaging.

Radical prostatectomy specimens used in this study were archived in IRB-approved genitourinary biorepositories at the University of Washington and University of Pennsylvania.

*B.  OTLS imaging and post processing*

Tri-labeled 100-μm-thick tissue curls (training data) and H&E-analog-stained 3-mm diameter prostate punches (clinical feasibility data) were imaged using a 4[th]-generation hybrid open-top light-sheet (OTLS) microscope [7], [23]. For the tri-labeled tissue curls, imaging was performed using three different excitation wavelengths (Cobolt Skyra, Hubner Photonics, Kasel, Germany): 488 nm for Alexa Fluor 488 NHS Ester, 561 nm for SYTO 85, and 638 nm for antibodies targeting PGP9.5 or CD31. For the prostate punch biopsies (clinical study data), excitation wavelengths of 561 nm and 638 nm were used for eosin and TO-PRO3. The acquired 3D volumes were stitched and fused using the BigStitcher plugin for ImageJ [37], followed by 2X down sampling in each dimension, resulting in a final sampling pitch of 0.527 μm/pixel in all three dimensions. The 3D volumes were then intensity-normalized (laterally and in depth) using a previously published post-processing method [24].

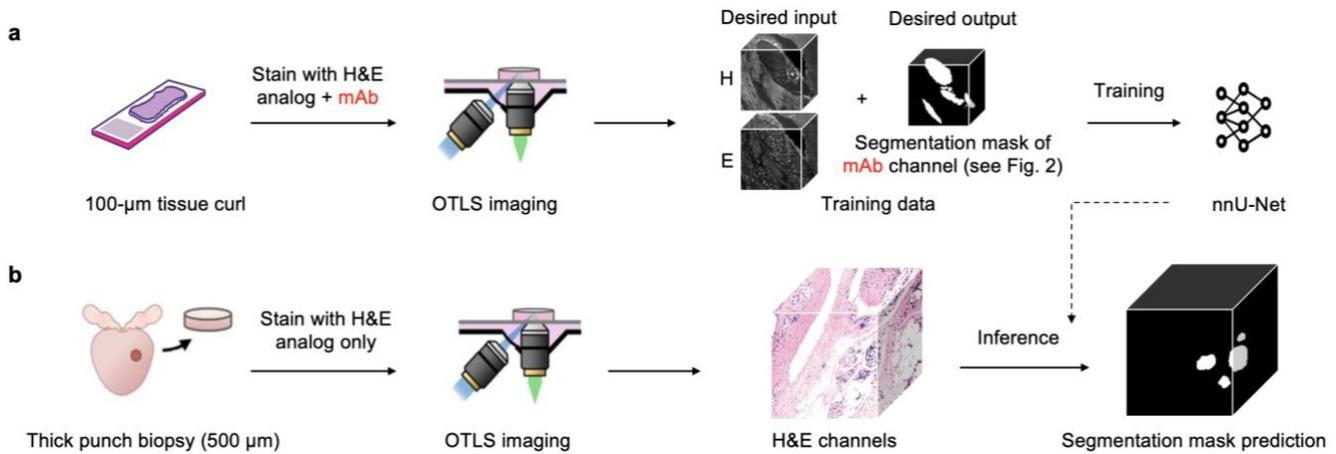

**Fig 1. Simplified workflow for training an nnU-Net segmentation model.** (a) The deep-learning (nnU-Net) segmentation model is trained with 3D pathology datasets of 100-micron thick tissues tri-labeled with a fluorescent analog of H&E plus a targeted monoclonal antibody (mAb). The fluorescent mAb channel is used to generate a ground-truth 3D segmentation mask, which is the desired output of the nnU-Net model. (b) Once trained, the nnU-Net model can generate 3D segmentation masks based on 3D pathology datasets of larger/thicker clinical specimens that are labeled with only H&E-analog fluorophores, which are inexpensive and can rapidly diffuse in thick 3D tissues (unlike mAbs).

*C. Segmentation pipeline*

*1) nnU-Net model*

We employed the nnU-Net framework to train a 3D segmentation model that predicts binary masks of nerve and vessel structures directly from H&E-analog fluorescence images. nnU-Net is a self-configuring deep learning pipeline built upon the U-Net architecture [38] and automatically adapts its preprocessing, network architecture, training, and post-processing steps to the input dataset [39], [40]. The model utilizes a combination of Dice loss and cross-entropy loss during training. Optimization was performed using Stochastic Gradient Descent with Nesterov momentum, and a poly learning rate schedule was applied to stabilize training and avoid gradient explosion.

*2) Training data preparation and ground truth generation*

For model training, regions of interest (ROIs) were extracted from the tri-labeled prostate datasets. Each ROI had a size of $512 \times 512 \times 50$ voxels at a resolution of 0.527 μm/pixel. A nnU-Net model was trained to predict the binary segmentation masks for nerves (PGP9.5-labeled) and vessels (CD31-labeled) using only the H&E-analog fluorescence channels as inputs (cytoplasmic and nuclear channels). As a preliminary study, we trained two models: a nerve segmentation model using 520 ROIs (PGP9.5-labeled), and a vessel segmentation model using 504 ROIs (CD31-labeled).

The PGP9.5 and CD31 immunolabeling channels were used to generate 3D segmentation masks that served as ground-truth training data for nerves and vessels, respectively. The immunolabeling channels were processed with computer vision techniques to create a binary mask of the corresponding labeled structures. Due to differences in staining quality, distinct strategies were used to generate ground-truth segmentation masks for nerves and vessels. The PGP9.5 channel showed strong and uniform staining, allowing fully automated 3D thresholding-based generation of ground truth nerve masks. In contrast, the CD31 staining was weaker and more discontinuous as it labels only the thin endothelial cell layer of the vessel wall without labeling the vessel lumens. To ensure accurate generation of ground truth vessel masks, we used a computer-assisted manual segmentation strategy to create complete vessel masks. See below for details.

To generate ground truth nerve masks, the PGP9.5 immunolabeling channel ($512 \times 512 \times \sim 50$ voxels) was first processed by logarithmic intensity adjustment to ensure signal uniformity and was then smoothed using a 3D Gaussian filter. Otsu's thresholding was further applied to determine an adaptive threshold and produce a 3D binary mask. This mask was refined through 3D morphological dilation using a spherical structuring element (radius = 7 voxels), hole-filling along all three orthogonal planes to ensure structural continuity, and a 3D erosion operation (radius = 3 voxels) to ensure continuity and boundary smoothness. The resulting masks provided reproducible ground truth labels for training a nerve segmentation model (**Fig 1a**).

For ground truth vessel masks, we used Medical SAM2 [41], a 3D extension of the Segment Anything Model, to facilitate the manual annotation of the CD31 immunolabeling channel. Annotators labelled only the central plane of each region of interest, and Medical SAM2 extrapolated a full 3D segmentation within each tissue volume based on context-aware features across neighboring slices (**Fig 1b**). The resulting masks were refined through a simple 3D binary closing operation using a spherical structuring element (radius = 3 voxels).

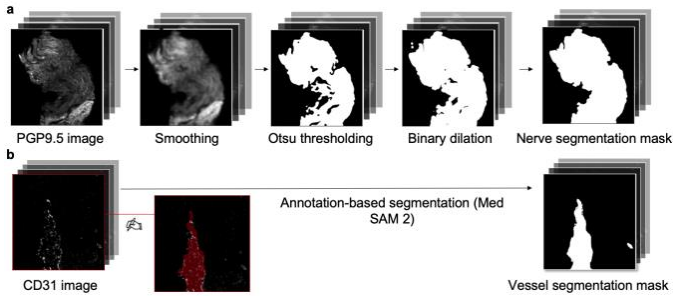

**Fig 1. Generating segmentation masks of mAb channels for nnU-Net training.** (a) Automated 3D thresholding-based computer vision method for generating ground truth segmentation masks of nerves. (b) Computer-assisted (Medical SAM2) annotation method for generating ground truth 3D segmentation masks of vessels.

*3) Clinical data preparation, inference, and volume assembly*

After imaging, 2X-downsampled 3D datasets of prostate punches are approximately 90 GB in size [24], with approximate dimensions of 4500 x 4500 x 1000 voxels (XYZ). To enable inference on GPUs with limited memory, the 3D datasets are divided into 25 overlapping sub-volumes (each 1024 × 1024 × 640 voxels, with 25% overlap in the XY dimension) and stored as .nii.gz files compatible with nnU-Net. The nuclei channel (TO-PRO3) and the Eosin channel serve as the two input channels for nnU-Net inference. After inference, output masks from each block are reassembled into a full-specimen segmentation mask. In overlapping regions, a voxel is assigned with a positive label for nerves/vessels (i.e., 1) if it is positive in any of the overlapping masks. Conversely, a voxel is assigned a background label (i.e., 0), indicating no nerves/vessels, if all contributing predictions classify it as background:

$$M_{final}(x, y, z) = \max(M_i(x, y, z)) \qquad (1)$$

where $M_i(x, y, z)$ denotes the predicted label at voxel $(x, y, z)$ from the $i^{th}$ overlapping block, and $M_{final}(x, y, z)$ is the resulting label in the final segmentation mask.

*4) Post-processing for nerve and vessel masks*

Following block-level inference of segmentation masks and stitching all blocks into a comprehensive whole-specimen 3D segmentation mask, post-processing steps were performed to refine the segmentation masks. First, the full 3D masks were cropped in the z-dimension to exclude non-tissue background and to retain only the tissue regions. To reduce computational time for downstream analysis while preserving a reasonable level of resolution, the 3D masks were down sampled by a factor of 4 in each dimension, resulting in final volumes of approximately 1024 x 1024 x 160 voxels (XYZ) at a sampling pitch of 2.1 μm/pixel. The down-sampled masks were then processed by a series of 3D morphological operations, including dilation, hole-filling, and erosion (as described in Section III-C-2), which collectively served to close gaps in the masks and consolidate fragmented structures such as partially stained nerves or vessel lumens. An additional correction step was applied only to the vessel segmentation masks to remove false-positive small vessels, such as those located within misclassified glandular regions. This was accomplished by comparing the vessel masks with a pre-existing glandular epithelium mask [36]. After an instance segmentation, any vessel *instance V* that had more than 10% of its volume overlapping with a gland region $G$ was discarded:

$$\frac{V \cap G}{V} > 0.1 \qquad (2)$$

Lastly, to focus our analysis on biologically relevant structures, small, isolated components were filtered out based on size. Only nerves and vessels above certain diameter thresholds (130 μm for nerves and 50 μm for vessels) were retained. These thresholds were determined based on published literature and empirical estimates for identifying clinically significant PNI and LVI [42]. To estimate the diameter of each segmented object, we applied principal component analysis (PCA) [43] to the voxels within each connected component in the 3D mask. Diameters of each isolated nerve/vessel fragment were calculated based on the minor axis perpendicular to the object's primary (major) axis.

*D. Comparing segmentation results with annotation ground truths*

To evaluate model performance, we selected 25 2D regions (2048 × 2048 pixels; 0.527 μm/pixel) from 7 prostate punch biopsies. Each region contained one or more nerve or vessel structures. A board-certified genitourinary pathologist manually annotated these structures based on false-colored H&E-analog fluorescence images, which closely resemble conventional H&E histology [44]. Annotations were performed using QuPath [45] and were exported as binary masks. Model predictions were then compared to these annotations to compute a Dice coefficient (F1 score), a standard metric for segmentation accuracy.

*E. Feature extraction and cancer region masks*

To characterize the features related to perineural invasion (PNI) and lymphovascular invasion (LVI), we performed quantitative spatial analysis of the proximity between cancer-enriched tissue regions and the segmented nerves or vessels. Cancer-enriched regions were defined using a recently developed pipeline, Synthetic Immunolabeling for Generative Heatmaps of Tumor (SIGHT) [36]. To extract PNI- and LVI-related features from these cancer masks and nerve/vessel segmentation masks, we developed two analysis strategies: (1) a 2D level-by-level analysis, designed to approximate conventional examination of histology slides, and (2) a 3D chunk-by-chunk analysis. Feature extraction was implemented in Python using *NumPy*, *SciPy*, and *scikit-image*.

*1) Level-by-level analysis*

Since the clinical diagnosis of PNI and LVI is based on 2D histology slides, features were first computed on a per-slice (2D) basis across all depth levels of the 3D pathology datasets. These features attempt to mirror conventional pathological definitions of PNI and LVI. For instance, PNI is classically defined by cancer surrounding at least 30% of the circumference of a nerve on a given 2D slide [26].

Level-by-level analysis starts with instance segmentation for each level along the z-axis of the stitched and post-processed nerve/vessel mask (**Fig 2a**). For each nerve or vessel instance $i$, we defined two annular zones: (1) an adjacent region $A_i$, extending ~32 μm from the boundary of the nerve/vessel (15 pixels), and (2) a distant region $B_i$, extending an additional ~32 μm from the first zone (15 to 30 pixels from the structure) as illustrated in **Fig 2b**. These dimensions (~32 μm) were optimized via computational ablation experiments (varying the dimension from 10 to 40 pixels, 21 to 84 μm). Within adjacent region $A_i$ and distant region $B_i$ of each nerve/vessel instance $i$, we quantified the cancer burden as the percentage of each region that overlaps with the binary cancer mask $C$:

$$Cancer\ burden\ in\ adjacent\ region = (A_i \cap C)/A_i \quad (3)$$
$$Cancer\ burden\ in\ distant\ region = (B_i \cap C)/B_i \quad (4)$$

From these, we derived the invasion score and gradient score for each nerve/vessel instance $i$:

$$Invasion\ score = \frac{(A_i \cap C)/A_i}{(B_i \cap C)/B_i} \quad (5)$$
$$Gradient\ score = \frac{(B_i \cap C)/B_i}{(A_i \cap C)/A_i} \quad (6)$$

A higher invasion score indicates that the cancer burden is greater near the nerve/vessel instance compared to that in the distant region. For each sample, summary statistics (mean, median, minimum, maximum, and standard deviation) were aggregated across all instances $i$ in each sample.

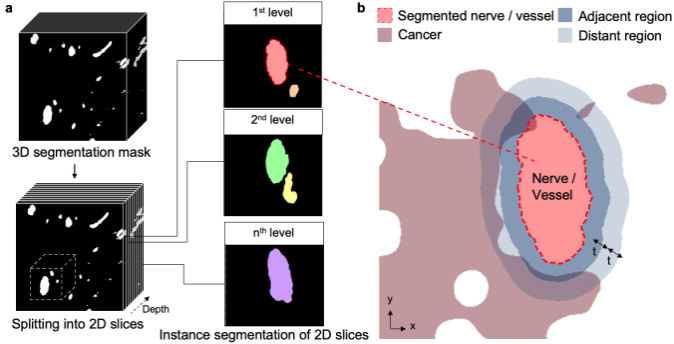

**Fig 2. Visualization of level-by-level annular analysis to extract PNI/LVI-related features.** (a) Stitched and post-processed segmentation masks are labelled with nerve/vessel instances. (b) Example of a nerve/vessel instance, adjacent annular region, distant annular region, and cancer mask. The annular thicknesses are $t$. The intersection between these regions is used to calculate features that are related to PNI/LVI.

*2) Chunk-by-chunk analysis*

While the level-by-level analysis mirrors how PNI and LVI are conventionally assessed in 2D histology, it does not capture the nerve/vessel-to-cancer spatial relationships in 3D. In 3D, nerves and vessels form a continuous interconnected network, and PNI/LVI can be seen in arbitrary directions. We thus developed a chunk-by-chunk analysis to quantify PNI/LVI-related features in 3D.

In this approach, the down-sampled segmentation mask was separated into overlapping 3D chunks of size 204 × 204 × 160 pixels (~ 428 x 428 x 336 μm). Chunks were extracted with a stride of half the chunk size (128 x 128 x 160 pixels), sliding across the tissue volume. The chunk size was optimized via computational ablation experiments (varying the dimension from 64 to 256 pixels, ~0.13 to 0.54 mm). Within each 3D chunk, we performed an analogous analysis to that described in Section III-E involving two regions: (1) an adjacent shell region $A$: extending 42 μm from the boundary of the nerve/vessel (~20 pixels), and (2) a distant shell region $B$, extending an additional 42 μm from the first zone (~20 to 40 pixels from the structure). Unlike the instance-based 2D analysis, these shell regions were defined globally within each chunk and could include multiple disconnected nerve or vessel regions. Similarly, we computed the cancer burden, and derived invasion and gradient scores using the same definitions as in Section III-E-1:

$$Cancer\ burden\ in\ adjacent\ region = (A \cap C)/A \quad (7)$$
$$Cancer\ burden\ in\ distant\ region = (B \cap C)/B \quad (8)$$
$$Invasion\ score = \frac{(A \cap C)/A}{(B \cap C)/B} \quad (9)$$
$$Gradient\ score = \frac{(B \cap C)/B}{(A \cap C)/A} \quad (10)$$

To obtain sample-level features, we aggregated the chunk-based features across the sample by calculating the mean, median, minimum, maximum, and standard deviation for each feature.

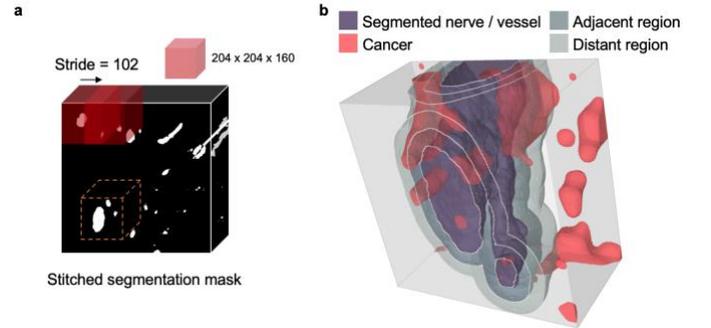

**Fig 3. Visualization of chunk-by-chunk shell analysis to extract PNI/LVI-related features.** (a) Stitched and post-processed segmentation masks are broken into chunks, with a stride of half the lateral dimension of the chunk size. (b) An example chunk is shown with a segmented nerve/vessel, an adjacent shell region, a distant shell region, and a cancer mask. The intersection between these regions is used to calculate features that are related to PNI/LVI.

*F. Feature evaluation*

To evaluate the extracted features, a LASSO-regularized logistic regression model was trained to perform binary classification of 5-year biochemical recurrence (BCR) outcomes. Patients who experienced BCR within 5 years post-radical prostatectomy (RP) are denoted as the "BCR" group, and all other patients are denoted as "non-BCR." BCR is defined as a rise in serum levels of prostate specific antigen (PSA) to 0.2 ng/mL at least 8 weeks after RP [46].

Within each training split, features were first filtered using an unsupervised feature selection approach to remove near-zero variance features and highly correlated features. The optimal LASSO regularization parameter ($C$) was selected via an inner 10-fold cross-validation loop, while model performance was evaluated using leave-one-out cross-validation (LOOCV) and quantified by the area under the receiver operating characteristic curve (AUC). All analyses were implemented in Python using *SciPy* and *scikit-learn*.

## IV. Results

Our results focus on validating the segmentation model for nerves and vessels, followed by clinical validation of PNI- and LVI-related features derived from both level-by-level and chunk-by-chunk methods. Across all experiments, LVI-related features were not significant predictors of clinical outcome as discussed in Section V.

### A. 3D nerve/vessel segmentation model performance

As mentioned, a 3D nnU-Net segmentation model was trained to automatically identify nerves and vessels in 3D pathology datasets of tissues stained with a fluorescent analog of H&E. The trained models were then applied to the clinical validation cohort. **Fig 4a** shows 3D renderings of predicted 3D segmentation masks for vessels and nerves. A false-colored H&E-mimicking image of a nerve or vessel (indicated by an arrow) is shown in the zoomed-in images shown in **Fig 4b**. Visual inspection confirms that both segmentation models successfully capture the morphology and distribution of nerve and vessel structures (**Fig 4c**). For quantitative validation, the overlap between ground truth annotations from a GU pathologist vs. model predictions is highlighted in white, with false negatives shown in magenta (pathologist's annotation positive, model negative) and false positives shown in cyan (pathologist's annotation negative, model positive). Segmentation performance was evaluated by comparing predicted masks with ground truth annotations for 25 2D regions (2048 × 2048 pixels; 0.527 µm/pixel) from 7 prostate punch biopsies (see Section III-D). The mean Dice similarity coefficient was $0.81 \pm 0.08$ for nerves and $0.71 \pm 0.09$ for vessels. For a video showing 3D renderings of the segmentation masks, please refer to the Supporting Materials.

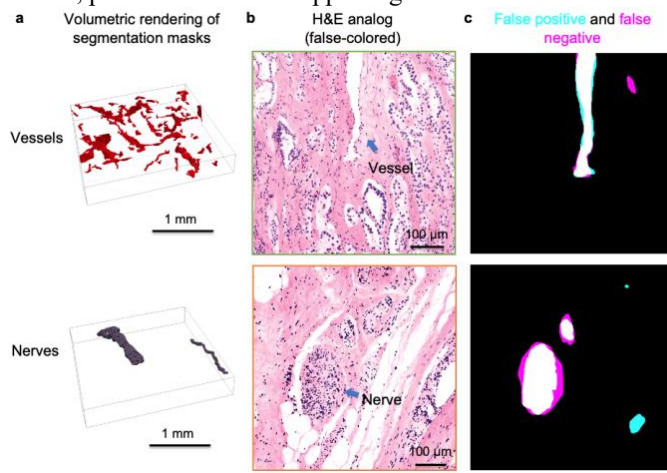

**Fig 4. Segmentation of nerves and vessels in 3D pathology datasets of prostate cancer.** The example images are from prostate cancer 3D pathology datasets showing 3D segmentations of vessels (first row) and nerves (second row). (a) 3D rendering of nerve/vessel segmentations in a 3-mm punch biopsy. (b) False-colored H&E-analog images of a zoomed-in region containing a nerve/vessel (indicated by an arrow). (c) Comparison between nnU-Net-generated segmentation masks and pathologist's ground-truth annotation. Agreement is shown in white, with false negatives shown in magenta (pathologist positive, model negative) and false positives shown in cyan (model positive, pathologist negative).

### B. Clinical validation result: correlating extracted features to BCR outcome

#### 1) Evaluation of level-by-level PNI-related features

We first validated PNI features derived from a level-by-level annular analysis (Section III-E). Segmentation was performed on 120 prostate specimens, of which 45 contained sufficient nerve structures for downstream feature extraction and biochemical recurrence (BCR) outcome prediction. For each nerve instance, PNI-related features were computed across all 2D tissue levels. To evaluate the effect of annulus size, the annular radius was systematically ablated from ~20 µm to ~80 µm, with a radius of ~30 µm yielding the most predictive features.

Extracted features were used to train a logistic regression model with LASSO regularization and leave-one-out cross-validation (LOOCV) (see Section III-F). As shown in **Figure 6a**, level-by-level features achieved an AUC of 0.71 (95% CI: 0.53–0.86) for distinguishing between high-risk vs. low-risk patients based on their 5-year BCR outcomes. To compare with standard clinical practice, where only three cross-sectional tissue levels (25 µm apart) are examined, we also developed risk-stratification models using features extracted from three depth levels (25 µm apart) within the 3D pathology datasets. Limiting our analysis to only three cross-sectional levels reduced the AUC to 0.52 (95% CI: 0.32–0.71), indicating that aggregating features across all depth levels (enabled by 3D pathology) better captures the spatial relationship between nerves and cancer-enriched regions and thus improves predictive performance.

#### 2) Evaluation of chunk-by-chunk PNI-related features

We next implemented the chunk-based shell analysis (Section III-E). This analysis was again restricted to the 45 patients with sufficient nerve content. The prostate volumes were partitioned into overlapping 3D chunks of 204 × 204 × 160 pixels (~0.4 × 0.4 × 0.33 mm), with a stride equal to half the chunk size (102 pixels). The shell radius was fixed at ~40 µm, close to the optimal parameter identified in the level-by-level study. For each chunk, adjacent vs. distant cancer burden metrics were calculated, allowing for a calculation of invasion and gradient scores.

Chunk-based features were used to train a LASSO-regularized model and evaluated using LOOCV. As shown in **Figure 6b**, PNI-related features from the chunk-based analysis demonstrate only modest prognostic value (AUC = 0.61), which was lower than that achieved by the level-by-level approach. The reduced performance might reflect dilution of local invasion signals across multiple small nerve fragments, as well as inclusion of irrelevant regions within chunks, which may limit discriminative power.

#### 3) Comparison between PNI-related feature and glandular features

Finally, we compared the predictive performance of PNI-related features and 3D glandular features. Glandular features were extracted as previously described [36], including measurements of glandular and luminal curvature, average distance to center of mass, and average 3D surface curvature, which have previously demonstrated prognostic value for prostate cancer.

Logistic regression models with LASSO regularization and LOOCV were trained using different feature sets. Models trained on PNI-related features alone achieved an AUC of 0.71 (95% CI: 0.53–0.86), while models using glandular features alone achieved an AUC of 0.64 (95% CI: 0.53,0.75) (**Fig 5c**).

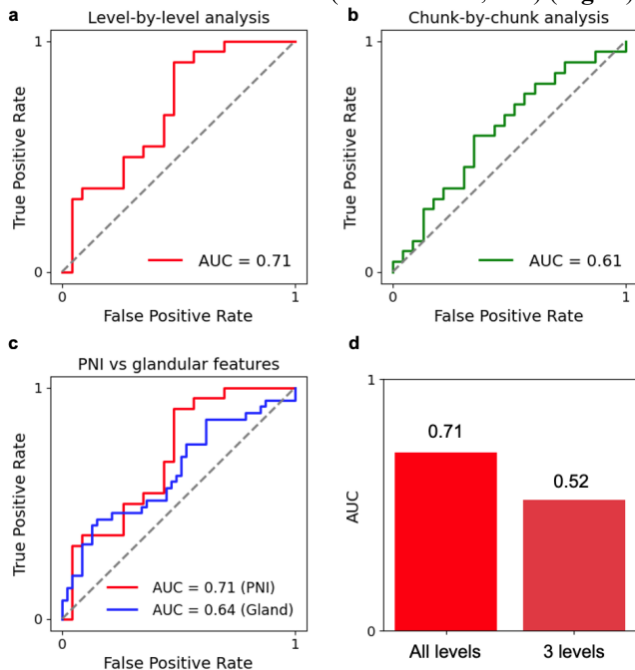

**Fig 5**. **Prognostic evaluation of PNI-related features in 3D pathology datasets.** Machine classifiers were trained to predict 5-year biochemical recurrence (BCR) outcomes based on PNI-related features from 45 patients. Prognostic performance is shown for models based on features derived from: (a) level-by-level annular analysis, (b) chunk-by-chunk shell analysis, and (c) glandular features compared with PNI-related features. PNI-related features are moderately prognostic (0.71 for level-by-level analysis and 0.61 by chunk-by-chunk analysis, n = 45). PNI-related features are superior to glandular features (AUC = 0.64, n = 45). (d) Performance improves when aggregating across all levels vs. only 3 levels.

## V. Discussion

In this work, we developed and validated a computational pipeline for extracting novel features to assess perineural invasion (PNI) and lymphovascular invasion (LVI) in 3D pathology datasets. These datasets were generated from prostate tissue punch biopsies imaged at sub-micron resolution using open-top light-sheet (OTLS) microscopy, enabling volumetric reconstructions of glandular, neural, and vascular structures. By leveraging our tri-labeling protocol, in which tissues were stained with a fluorescent analog of H&E as well as an antibody that targets either nerves (PGP9.5) or vessels (CD31), we trained a nnU-Net deep learning model to generate 3D segmentation masks of nerves and vessels in prostate tissues. We then implemented 2D level-by-level annular analyses and 3D shell-based chunk analyses to capture histomorphological insights into how cancer-enriched regions interact with nerves and vessels in prostate tissue. Together, these techniques mimic conventional pathological definitions of PNI and LVI for 2D histology, while also extending the analysis into three dimensions through chunk-based shell analysis. In summary, we have developed new methods to quantify spatial patterns of tumor–nerve and tumor–vessel interactions in large-scale 3D pathology datasets.

Preliminary clinical-feasibility studies reveal that novel PNI-related features provide moderate predictive performance. Limitations is predictive performance can be attributed to several factors. First, the size of the clinical cohort available for analysis was relatively small (n=45) and skewed toward lower-grade patients (mostly Gleason grade groups 1 and 2). Second, although PNI has long been reported to correlate with adverse clinical outcomes, its mechanism and prognostic value remain complex. Not all patients with aggressive disease exhibit PNI and prostate cancer can progress and disseminate through multiple biological pathways. Despite these limitations, the PNI-related features demonstrated predictive performance comparable to that of established gland-derived morphometric features. Further studies with larger and more diverse cohorts are needed to determine whether PNI-related features can synergize with additional histomorphometric and microenvironmental features to support comprehensive and robust predictive models.

The cancer masks generated by the SIGHT pipeline provide gland-level resolution but are coarse compared with the cellular-level precision required to directly observe individual tumor cells infiltrating along nerves or vessels. Consequently, extracted features may not fully capture the fine-scale mechanics of PNI and LVI. In addition, SIGHT was not trained to detect tumor cells within lymphovascular lumen spaces (true instances of LVI). Finally, LVI is a relatively rare event in prostate cancer and is less consistently observed than PNI in clinical practice (for prostate cancer). Therefore, it is unsurprising that the LVI-related features in this study are not as prognostic as PNI-related features for prostate cancer.

From a technical perspective, implementing this framework required careful consideration of the scale and complexity of the data. Whole prostate tissue blocks digitized at sub-micron resolution generate datasets of several hundred gigabytes, necessitating strategies such as downsampling for feature extraction and block-wise chunking for both segmentation inference and feature analysis. These design choices enabled the workflow to be executed within the constraints of GPU memory and computational resources. The choice to analyze local 3D chunks and local instances of nerves or vessels, rather than attempt full-volume feature extraction, also reflects the biological nature of the problem: PNI and LVI are relatively rare, localized events. Without localized analysis, any signal associated with these events would be diluted when aggregated across the entirety of a nerve or vessel network.

In summary, our study provides the first end-to-end computational pipeline for quantifying PNI and LVI in 3D pathology datasets. This involved developing a method to generate 3D segmentation masks of nerves and vessels in 3D pathology datasets and then developing new methods to extract novel features that capture both 2D and 3D spatial relationships between cancer-enriched regions and nerves/vessels. Finally, we provide proof-of-concept results showing that PNI- and LVI- related features extracted from 3D pathology datasets can potentially improve risk stratification and oncologic decisions.

VI. REFERENCES

[1] R. L. Siegel, A. N. Giaquinto, and A. Jemal, "Cancer statistics, 2024," *CA A Cancer J Clinicians*, vol. 74, no. 1, pp. 12–49, Jan. 2024.
[2] J. I. Epstein, M. B. Amin, S. W. Fine, et al., "The 2019 Genitourinary Pathology Society (GUPS) White Paper on Contemporary Grading of Prostate Cancer," *Arch Pathol Lab Med*, vol. 145, no. 4, pp. 461–493, Apr. 2021.
[3] J. I. Epstein, M. J. Zelefsky, D. D. Sjoberg, et al., "A Contemporary Prostate Cancer Grading System: A Validated Alternative to the Gleason Score," *Eur Urol*, vol. 69, no. 3, pp. 428–435, Mar. 2016.
[4] D. F. Gleason, "Classification of prostatic carcinomas," *Cancer Chemother Rep*, vol. 50, no. 3, pp. 125–128, Mar. 1966.
[5] G. Gao, R. Yan, A. H. Song, et al., "Deep-learning triage of 3D pathology datasets for comprehensive and efficient pathologist assessments," *bioRxiv*, p. 2025.07.20.665804, Jan. 2025.
[6] C. Koyuncu, A. Janowczyk, X. Farre, et al., "Visual Assessment of 2-Dimensional Levels Within 3-Dimensional Pathology Data Sets of Prostate Needle Biopsies Reveals Substantial Spatial Heterogeneity," *Lab Invest*, vol. 103, no. 12, p. 100265, Dec. 2023.
[7] A. K. Glaser, N. P. Reder, Y. Chen, et al., "Light-sheet microscopy for slide-free non-destructive pathology of large clinical specimens," *Nat Biomed Eng*, vol. 1, no. 7, p. 0084, June 2017.
[8] J. T. Liu, S. S. Chow, R. Colling, et al., "Engineering the future of 3D pathology," *The Journal of Pathology CR*, vol. 10, no. 1, p. e347, Jan. 2024.
[9] N. P. Reder, A. K. Glaser, E. F. McCarty, et al., "Open-Top Light-Sheet Microscopy Image Atlas of Prostate Core Needle Biopsies," *Archives of Pathology & Laboratory Medicine*, vol. 143, no. 9, pp. 1069–1075, Sept. 2019.
[10] L. Egevad, A. S. Ahmad, F. Algaba, et al., "Standardization of Gleason grading among 337 European pathologists," *Histopathology*, vol. 62, no. 2, pp. 247–256, Jan. 2013.
[11] C. J. Kane, S. E. Eggener, A. W. Shindel, et al., "Variability in Outcomes for Patients with Intermediate-risk Prostate Cancer (Gleason Score 7, International Society of Urological Pathology Gleason Group 2-3) and Implications for Risk Stratification: A Systematic Review," *Eur Urol Focus*, vol. 3, no. 4–5, pp. 487–497, Oct. 2017.
[12] T. A. Ozkan, A. T. Eruyar, O. O. Cebeci, et al., "Interobserver variability in Gleason histological grading of prostate cancer," *Scand J Urol*, vol. 50, no. 6, pp. 420–424, Dec. 2016.
[13] M. C. Haffner, A. M. De Marzo, S. Yegnasubramanian, et al., "Diagnostic challenges of clonal heterogeneity in prostate cancer," *J Clin Oncol*, vol. 33, no. 7, pp. e38-40, Mar. 2015.
[14] A. U. Frey, J. Sønksen, and M. Fode, "Neglected side effects after radical prostatectomy: a systematic review," *J Sex Med*, vol. 11, no. 2, pp. 374–385, Feb. 2014.
[15] N. Renier, Z. Wu, D. J. Simon, et al., "iDISCO: A Simple, Rapid Method to Immunolabel Large Tissue Samples for Volume Imaging," *Cell*, vol. 159, no. 4, pp. 896–910, Nov. 2014.
[16] E. A. Susaki, K. Tainaka, D. Perrin, et al., "Advanced CUBIC protocols for whole-brain and whole-body clearing and imaging," *Nat Protoc*, vol. 10, no. 11, pp. 1709–1727, Nov. 2015.
[17] R. Serafin, C. Koyuncu, W. Xie, et al., "Nondestructive 3D pathology with analysis of nuclear features for prostate cancer risk assessment," *The Journal of Pathology*, vol. 260, no. 4, pp. 390–401, Aug. 2023.
[18] A. H. Song, M. Williams, D. F. K. Williamson, et al., "Analysis of 3D pathology samples using weakly supervised AI," *Cell*, vol. 187, no. 10, pp. 2502-2520.e17, May 2024.
[19] W. Xie, N. P. Reder, C. Koyuncu, et al., "Prostate Cancer Risk Stratification via Nondestructive 3D Pathology with Deep Learning–Assisted Gland Analysis," *Cancer Research*, vol. 82, no. 2, pp. 334–345, Jan. 2022.
[20] O. L. Katsamenis, M. Olding, J. A. Warner, et al., "X-ray Micro-Computed Tomography for Nondestructive Three-Dimensional (3D) X-ray Histology," *Am J Pathol*, vol. 189, no. 8, pp. 1608–1620, Aug. 2019.
[21] A. L. Kiemen, A. M. Braxton, M. P. Grahn, et al., "CODA: quantitative 3D reconstruction of large tissues at cellular resolution," *Nat Methods*, vol. 19, no. 11, pp. 1490–1499, Nov. 2022.
[22] J. Park, S.-J. Shin, G. Kim, et al., "Revealing 3D microanatomical structures of unlabeled thick cancer tissues using holotomography and virtual H&E staining," *Nat Commun*, vol. 16, no. 1, p. 4781, May 2025.
[23] L. A. Barner, A. K. Glaser, H. Huang, et al., "Multi-resolution open-top light-sheet microscopy to enable efficient 3D pathology workflows," *Biomed. Opt. Express*, vol. 11, no. 11, p. 6605, Nov. 2020.
[24] K. W. Bishop, L. A. Erion Barner, Q. Han, et al., "An end-to-end workflow for nondestructive 3D pathology," *Nat Protoc*, vol. 19, no. 4, pp. 1122–1148, Apr. 2024.
[25] J. T. C. Liu, A. K. Glaser, K. Bera, et al., "Harnessing non-destructive 3D pathology," *Nat Biomed Eng*, vol. 5, no. 3, pp. 203–218, Feb. 2021.
[26] S.-H. Chen, B.-Y. Zhang, B. Zhou, et al., "Perineural invasion of cancer: a complex crosstalk between cells and molecules in the perineural niche," *Am J Cancer Res*, vol. 9, no. 1, pp. 1–21, 2019.
[27] L. Cheng, T. D. Jones, H. Lin, et al., "Lymphovascular invasion is an independent prognostic factor in prostatic adenocarcinoma," *J Urol*, vol. 174, no. 6, pp. 2181–2185, Dec. 2005.
[28] P. Zareba, R. Flavin, M. Isikbay, et al., "Perineural Invasion and Risk of Lethal Prostate Cancer," *Cancer Epidemiology, Biomarkers & Prevention*, vol. 26, no. 5, pp. 719–726, May 2017.
[29] J. A. Cohn, P. P. Dangle, C. E. Wang, et al., "The prognostic significance of perineural invasion and race in men considering active surveillance," *BJU International*, vol. 114, no. 1, pp. 75–80, July 2014.
[30] D. J. Grignon, "Prostate cancer reporting and staging: needle biopsy and radical prostatectomy specimens," *Modern Pathology*, vol. 31, pp. 96–109, Jan. 2018.
[31] E. I. Harris, D. N. Lewin, H. L. Wang, et al., "Lymphovascular Invasion in Colorectal Cancer: An Interobserver Variability Study," *American Journal of Surgical Pathology*, vol. 32, no. 12, pp. 1816–1821, Dec. 2008.
[32] M. M. Fraz, S. A. Khurram, S. Graham, et al., "FABnet: feature attention-based network for simultaneous segmentation of microvessels and nerves in routine histology images of oral cancer," *Neural Comput & Applic*, vol. 32, no. 14, pp. 9915–9928, July 2020.
[33] K. Kartasalo, P. Ström, P. Ruusuvuori, et al., "Detection of perineural invasion in prostate needle biopsies with deep neural networks," *Virchows Arch*, vol. 481, no. 1, pp. 73–82, July 2022.
[34] M. Amgad, H. Elfandy, H. Hussein, et al., "Structured crowdsourcing enables convolutional segmentation of histology images," *Bioinformatics*, vol. 35, no. 18, pp. 3461–3467, Sept. 2019.
[35] M. I. Todorov, J. C. Paetzold, O. Schoppe, et al., "Machine learning analysis of whole mouse brain vasculature," *Nat Methods*, vol. 17, no. 4, pp. 442–449, Apr. 2020.
[36] R. B. Serafin, J. S. Lopez, S. Chow, et al., "Detection of prostate cancer in 3D pathology datasets via generative immunolabeling," *bioRxiv*, p. 2025.08.11.669726, Jan. 2025.
[37] D. Hörl, F. Rojas Rusak, F. Preusser, et al., "BigStitcher: reconstructing high-resolution image datasets of cleared and expanded samples," *Nat Methods*, vol. 16, no. 9, pp. 870–874, Sept. 2019.
[38] O. Ronneberger, P. Fischer, and T. Brox, "U-Net: Convolutional Networks for Biomedical Image Segmentation," in *Medical Image Computing and Computer-Assisted Intervention – MICCAI 2015*, vol. 9351, N. Navab, J. Hornegger, W. M. Wells, et al., Eds. Cham: Springer International Publishing, 2015, pp. 234–241.
[39] F. Isensee, P. F. Jaeger, S. A. A. Kohl, et al., "nnU-Net: a self-configuring method for deep learning-based biomedical image segmentation," *Nat Methods*, vol. 18, no. 2, pp. 203–211, Feb. 2021.
[40] R. Wang, S. S. L. Chow, R. B. Serafin, et al., "Direct three-dimensional segmentation of prostate glands with nnU-Net," *J. Biomed. Opt.*, vol. 29, no. 03, Mar. 2024.
[41] J. Zhu, A. Hamdi, Y. Qi, et al., "Medical SAM 2: Segment medical images as video via Segment Anything Model 2." arXiv, 04-Dec-2024.
[42] L. B. Schmitd, L. J. Beesley, N. Russo, et al., "Redefining Perineural Invasion: Integration of Biology With Clinical Outcome," *Neoplasia*, vol. 20, no. 7, pp. 657–667, July 2018.
[43] A. Maćkiewicz and W. Ratajczak, "Principal components analysis (PCA)," *Computers & Geosciences*, vol. 19, no. 3, pp. 303–342, Mar. 1993.
[44] R. Serafin, W. Xie, A. K. Glaser, et al., "FalseColor-Python: A rapid intensity-leveling and digital-staining package for fluorescence-based slide-free digital pathology," *PLoS ONE*, vol. 15, no. 10, p. e0233198, Oct. 2020.

[45] P. Bankhead, M. B. Loughrey, J. A. Fernández, et al., "QuPath: Open source software for digital pathology image analysis," *Sci Rep*, vol. 7, no. 1, p. 16878, Dec. 2017.

[46] S. Hawley, L. Fazli, J. K. McKenney, et al., "A model for the design and construction of a resource for the validation of prognostic prostate cancer biomarkers: the Canary Prostate Cancer Tissue Microarray," *Adv Anat Pathol*, vol. 20, no. 1, pp. 39–44, Jan. 2013.